\documentclass{article}

\usepackage{microtype}
\usepackage{graphicx}
\usepackage{subcaption}
\usepackage{booktabs}
\usepackage{multirow}
\usepackage{hyperref}
\usepackage{tikz}
\usetikzlibrary{arrows.meta,positioning,shapes.geometric}
\usepackage{pgfplots}
\pgfplotsset{compat=1.18}

\usepackage[accepted]{icml2026}

\makeatletter
\renewcommand{\ICML@appearing}{\textit{Accepted to the
ICML 2026 RLxF: Reinforcement Learning from World Feedback Workshop},
RLxF@ICML 2026, Seoul, South Korea.
Copyright 2026 by the author(s).}
\makeatother

\usepackage{amsmath}
\usepackage{amssymb}
\usepackage{mathtools}
\usepackage{amsthm}

\usepackage[capitalize,noabbrev]{cleveref}
\usepackage{pifont}

\usepackage{soul}
\definecolor{hlred}{RGB}{255,204,204}
\definecolor{hlgreen}{RGB}{204,255,204}

\theoremstyle{plain}

\theoremstyle{definition}

\icmltitlerunning{Closing the Feedback Loop: From Experience Extraction to Insight Governance}

\begin{document}

\twocolumn[
  \icmltitle{Closing the Feedback Loop: From Experience Extraction to\\Insight Governance in Verbal Reinforcement Learning}

  \begin{icmlauthorlist}
    \icmlauthor{Yanwei Cui}{aff1}
    \icmlauthor{Xing Zhang}{aff1}
    \icmlauthor{Yulong Zhang}{aff2}
    \icmlauthor{Li Shao}{aff2}
    \icmlauthor{Xiaofeng Shi}{aff3}
    \icmlauthor{Guanghui Wang}{aff1}
    \icmlauthor{Peiyang He}{aff1}
  \end{icmlauthorlist}

  \icmlaffiliation{aff1}{AWS Generative AI Innovation Center}
  \icmlaffiliation{aff2}{Amazon Web Services (AWS)}
  \icmlaffiliation{aff3}{BingX Group Limited}

  \icmlcorrespondingauthor{Peiyang He}{peiyan@amazon.com}

  \icmlkeywords{LLM Agents, World Feedback, Knowledge Governance, Verbal Reinforcement Learning, Agent Memory}

  \vskip 0.3in
]

\printAffiliationsAndNotice{}

\begin{abstract}
Training-free verbal reinforcement learning enables LLM agents to learn from world feedback---objective signals such as dynamic task outcomes, market returns, or demand forecasts---by extracting verbal rules from experience and injecting them as context, updating the agent's behavior without parameter changes.
However, in non-stationary environments these agents face a retention-forgetting dilemma: retaining stale insights causes negative transfer, while discarding them causes catastrophic forgetting when conditions recur.
We identify four requirements for navigating this dilemma---outcome-driven evaluation, persistent structured evidence, non-monotonic knowledge lifecycle, and compositional governance---and show that existing methods invest heavily in experience extraction while underinvesting in insight governance.
We propose a three-layer architecture---rules, evidence, and skills---connected by a feedback-driven curation loop that closes the governance gap.
Rules capture distilled experience from world outcomes; evidence logs track each rule's reliability across episodes; skills govern which rules to apply, how to resolve conflicts, and when to abstain.
On financial forecasting as a case study, where world feedback is naturally abundant, noisy, and non-stationary, we show that the same accumulated experience either degrades performance below the zero-shot baseline or dramatically improves accuracy and risk-adjusted returns, depending on whether the curation loop is present.
\end{abstract}

\section{Introduction}
\label{sec:intro}

LLM agents increasingly operate in domains where \emph{world feedback}---objective signals arising from real-world interactions such as dynamic task outcomes, market returns, or demand forecasts---arrives after the agent acts.
A growing body of work treats this world feedback as a first-class learning signal, enabling agents to improve without gradient updates by extracting verbal rules from experience and injecting them as context~\citep{shinn2023reflexion,zhao2023expel,cai2025trainingfreegrpo,allard2026erl}.
This paradigm---\emph{verbal reinforcement learning from world feedback}---updates the agent's context rather than its parameters, offering an interpretable and modular alternative to fine-tuning.

But a fundamental problem is underexplored: \textbf{in non-stationary environments, accumulated experience can hurt as much as it helps.}
Rules that worked under one regime may fail when conditions shift---and most real-world feedback environments are non-stationary.
An agent that stores everything drowns in contradictory context; an agent that discards failures forgets lessons it will need when conditions recur.
We call this the \textbf{retention-forgetting dilemma} and argue it is the central design challenge for any agent learning from non-stationary world feedback.

We identify four requirements that a learning system must satisfy to navigate this dilemma (\cref{sec:requirements}): outcome-driven evaluation (R1), persistent structured evidence (R2), non-monotonic knowledge lifecycle (R3), and compositional governance (R4).
Examining recent training-free methods (\cref{sec:gaps}), we find that while individual requirements are increasingly addressed, no existing approach satisfies all four.
This finding is consistent with concurrent empirical evidence from SkillsBench~\citep{li2026skillsbench}, which shows that for static procedural-knowledge packages, curated skills substantially improve agent performance while self-generated skills do not---locating skill curation as a design axis that meaningfully shapes outcomes.

We propose closing the loop with a three-layer architecture (\cref{sec:approach}): \emph{rules} capture distilled experience, \emph{evidence} logs track each rule's reliability across episodes, and \emph{skills} govern which rules to apply, how to resolve conflicts, and when to abstain.
Three curation roles---critic, proposer, and curator---connect these layers through a feedback-driven loop where world outcomes drive knowledge lifecycle decisions.
Each layer is motivated by a failure mode of the layer below: rules alone cannot tell the agent which ones to trust; per-rule evidence alone cannot handle composition; only skills operating over evidence can provide principled governance.

We validate on financial forecasting (\cref{sec:experiments}), where world feedback is naturally abundant, objective, noisy, delayed, and non-stationary.
The results demonstrate a striking pattern: the same accumulated experience either degrades or dramatically improves performance, depending solely on which requirements are satisfied.

Our contributions are: (1) the retention-forgetting dilemma as a framing for the central challenge of verbal RL from non-stationary world feedback; (2) four requirements (R1--R4) that characterize the gap between experience extraction and insight governance in existing methods; (3) a three-layer architecture with a feedback-driven curation loop designed to close this gap; and (4) empirical evidence that governance---not the quantity of accumulated experience---determines whether an agent improves or degrades.

\section{The Problem: Learning from World Feedback}
\label{sec:problem}

\subsection{The Retention-Forgetting Dilemma}
\label{sec:dilemma}

When an agent accumulates experience from world feedback in a non-stationary environment, it faces a fundamental tension:

\begin{itemize}
  \item \textbf{Retain everything} $\rightarrow$ the agent's context fills with stale and contradictory rules. The wrong rule fires at the wrong time, producing confident but wrong outputs. Performance degrades below zero-shot---experience actively hurts.
  \item \textbf{Discard what fails} $\rightarrow$ when conditions recur (and in non-stationary environments, they do), the agent has no memory of what worked before. It re-learns from scratch, paying the same cost again.
\end{itemize}

This dilemma arises wherever world feedback is non-stationary: financial markets exhibit regime shifts, robotic control environments change through wear and perturbation, and demand patterns drift with seasons and policy changes.
The question is not whether accumulated experience will eventually become stale, but how the agent manages it when it does.

\subsection{Requirements for Effective Learning}
\label{sec:requirements}

We identify four requirements that any system must satisfy to navigate the retention-forgetting dilemma:

\paragraph{R1. Outcome-driven evaluation.}
The system must systematically evaluate whether stored knowledge actually helped, based on observed outcomes---not just whether the task succeeded, but \emph{how} the knowledge affected the agent's reasoning.
Without this, the agent cannot distinguish useful knowledge from noise.
SkillsBench~\citep{li2026skillsbench} reports that curated skills substantially improve agent performance while self-generated skills do not, indicating that the vetting of procedural knowledge matters for static skill packages. Our R1 asks the complementary dynamic-setting question: how to vet rules continuously as world outcomes arrive.

\paragraph{R2. Persistent, structured evidence.}
Evaluation signals must accumulate across episodes and remain linked to the specific knowledge they concern.
A single episode is too noisy to draw conclusions; cross-episode evidence is what separates signal from noise.
When knowledge is modified or retired, the evidence trail must survive---otherwise the system loses the basis for future decisions.
Hindsight~\citep{latimer2025hindsight} demonstrates the value of tracking belief strength through its Opinion Network, where confidence scores evolve as new evidence arrives.
However, scalar confidence scores discard the structured evidence trail: when a belief's confidence drops from 0.85 to 0.55, the system retains no record of \emph{which facts} caused the change or under \emph{what conditions}.

\paragraph{R3. Non-monotonic knowledge lifecycle.}
The system must be able to both add and deactivate knowledge.
Critically, deactivation should not mean deletion---deprecated knowledge and its evidence should be preserved so the system does not forget what it learned.
This resolves the dilemma: deactivated rules cause no negative transfer, but their evidence prevents catastrophic forgetting.
The AGM belief revision framework~\citep{alchouron1985logic} formalizes why this matters: the Relevance postulate (minimal change) and Core-Retainment (no unjustified deletion) provide mathematical guarantees that knowledge removal preserves maximal information.
Recent systems like Kumiho~\citep{park2026kumiho} demonstrate that these formal guarantees are operationally feasible for agent memory graphs.

\paragraph{R4. Compositional governance.}
Individual rules interact: they may conflict, reinforce, or apply only in certain conditions.
The system needs a higher-order mechanism---what we call \emph{skills}---that governs which rules to apply, how to resolve conflicts, and when to abstain.
Without this, the agent is at the mercy of whichever rule happens to match most closely.
SkillsBench~\citep{li2026skillsbench} reports that comprehensive skill sets can degrade agent performance while focused skill sets improve it, and explicitly identifies skills composition as an open problem.

\subsection{Gaps in Existing Approaches}
\label{sec:gaps}

Among the training-free verbal learning methods we reviewed (\cref{tab:gaps}), individual requirements are increasingly addressed, but a unified solution remains elusive.

\begin{table}[t]
  \caption{Requirements satisfied by training-free verbal reinforcement learning approaches. Existing methods invest heavily in experience extraction but underinvest in insight governance.}
  \label{tab:gaps}
  \begin{center}
    \resizebox{\columnwidth}{!}{%
    \begin{small}
        \begin{tabular}{lcccc}
          \toprule
          Approach & R1 & R2 & R3 & R4 \\
          \midrule
          Reflective accumulation & Partial & \ding{55} & \ding{55} & \ding{55} \\
          {\scriptsize \citep{shinn2023reflexion,allard2026erl}} & {\scriptsize (trajectory)} & & & \\
          \addlinespace
          Reflective refinement & Partial & \ding{55} & Partial & \ding{55} \\
          {\scriptsize \citep{zhao2023expel,cai2025trainingfreegrpo}} & {\scriptsize (scalar)} & {\scriptsize (lost on mod.)} & {\scriptsize (in-place)} & \\
          \addlinespace
          Trajectory-informed tips & \ding{51} & \ding{55} & Partial & Partial \\
          {\scriptsize \citep{fang2025trajectory}} & {\scriptsize (causal attr.)} & {\scriptsize (no cross-episode)} & {\scriptsize (consolidation)} & {\scriptsize (LLM-guided)} \\
          \addlinespace
          Meta-MDP experience library & \ding{51} & \ding{55} & Partial & Partial \\
          {\scriptsize \citep{cai2025flex}} & {\scriptsize (semantic critic)} & {\scriptsize (merge erases)} & {\scriptsize (preserved failures)} & {\scriptsize (3-level hier.)} \\
          \midrule
          \textbf{Ours} & \ding{51} & \ding{51} & \ding{51} & \ding{51} \\
          & {\scriptsize (reasoning)} & {\scriptsize (cross-batch)} & {\scriptsize (deprecation)} & {\scriptsize (evolving)} \\
          \bottomrule
        \end{tabular}
    \end{small}
    }%
  \end{center}
  \vskip -0.1in
\end{table}

\textbf{Reflective accumulation}~\citep{shinn2023reflexion,allard2026erl} extracts verbal feedback from errors after each episode and appends it to the agent's context.
Reflection is triggered by task outcomes (partial R1 at the trajectory level), but stored reflections are never subsequently evaluated against later outcomes---all accumulated experience is kept and treated equally, regardless of whether any given reflection helped or hurt downstream.

\textbf{Reflective refinement}~\citep{zhao2023expel,cai2025trainingfreegrpo} extends reflective accumulation with importance scoring and in-place rule modification.
This satisfies R1 partially (scalar evaluation signals exist) and R3 partially (rules are modified rather than only added).
However, in-place modification destroys evidence: when a rule is rewritten, all previously accumulated evaluation signals are invalidated, requiring costly re-evaluation to rebuild confidence.

\textbf{Trajectory-informed tips}~\citep{fang2025trajectory} introduces automated causal attribution over trajectories, satisfying R1 through principled outcome-driven evaluation.
Tips carry structured provenance and are consolidated at storage time through deduplication, conflict resolution, and merging (partial R3); at retrieval, an LLM-guided selector filters by task context and priority (partial R4).
However, a stored tip never accumulates additional evidence from subsequent episodes, and the system merges or overwrites tips rather than deprecating them---leaving R2 unaddressed and R3 handled only via modification.

\textbf{Meta-MDP experience library}~\citep{cai2025flex} casts training-free learning as a Meta-MDP with two-level evaluation---a semantic critic at the trajectory level and a ground-truth reward at the library level---cleanly satisfying R1.
The library is split into \emph{golden} (distilled successes) and \emph{warning} (failure lessons) zones, explicitly preserving failure knowledge (partial R3); but zone assignment is fixed at intake, so evidence-driven demotion is absent.
R2 is unaddressed: the updater merges semantically similar entries into one record, erasing which source trajectories contributed and under what conditions.
Retrieval is three-level hierarchical top-$k$ (partial R4), without conflict resolution or abstention.

Complementary benchmark evidence reinforces that the governance gap is real. SkillsBench~\citep{li2026skillsbench} is a static evaluation---skills are injected once per task with no cross-episode feedback, so it is not itself a verbal-RL method---but it finds that \emph{curated} skill packages yield $+16.2$pp pass-rate gains while \emph{self-generated} skills give $-1.3$pp on average, and that focused skills outperform comprehensive ones. The first result shows that curation quality, not skill quantity, drives gains; the second shows that skill composition is a real design axis. Both motivate R1 (quality-checking) and R4 (composition), and SkillsBench explicitly identifies lifecycle and compositional governance as open problems.

The key pattern across existing verbal-RL methods: they invest heavily in \textbf{extraction}---how to produce good rules from experience---but underinvest in \textbf{governance}---how to manage rules once they exist.
R1 (evaluation) is the most developed, from proximity-based credit assignment to deterministic verification.
But R2 (persistent evidence), R3 (non-monotonic lifecycle), and R4 (compositional governance) remain largely unsatisfied.

\subsection{Advanced Agent Memory Systems}
\label{sec:memory}

A parallel line of work develops the memory infrastructure that any learning agent requires to store, retrieve, and update knowledge.
These systems provide storage and retrieval primitives that our architecture assumes; our contribution is the feedback-driven curation loop that sits on top.

Hindsight~\citep{latimer2025hindsight} is the most relevant architecture: four epistemic networks (world, experience, opinion, observation) with three operations---\emph{retain, recall, reflect}---and an Opinion Network whose beliefs carry evolving confidence scores.
The paradigm provides the right structure for memory management, but \emph{reflect} updates beliefs by factual consistency rather than outcome-driven evaluation, and scalar confidence discards the structured evidence trail---when a score drops, the system cannot reconstruct why.

IMPACT-CYCLE~\citep{kong2026impactcycle} demonstrates how provenance logs and dependency-closure correction can maintain persistent evidence (R2) with localized non-monotonic updates (R3) in a multi-agent supervisory system for long-video semantic memory.
Each claim carries a dependency graph, and corrections propagate only to structurally dependent claims.
However, IMPACT-CYCLE corrects factual claims within a single session rather than managing predictive rules across episodes.

At the formal level, the AGM belief revision framework~\citep{alchouron1985logic} provides mathematical guarantees for knowledge lifecycle.
The Relevance postulate ensures minimal change during revision; Core-Retainment prevents unjustified deletion.
Recent systems like Kumiho~\citep{park2026kumiho} demonstrate the operational feasibility of these guarantees for agent memory, implementing AGM-compliant belief revision over graph-native memory architectures.
MemGPT~\citep{packer2023memgpt} pioneered the concept of virtual context management for LLM agents, establishing the tiered memory architecture that many subsequent systems build upon.

These memory systems manage \emph{what the agent remembers}.
Our curation loop manages \emph{what the agent should trust}---connecting world outcomes back to stored knowledge through evaluation, evidence, principled deprecation, and compositional governance.

\section{Approach: Feedback-Driven Curation}
\label{sec:approach}

\subsection{Architecture: Three Layers, One Loop}
\label{sec:layers}

We propose three layers, each motivated by a failure mode of the layer below (\cref{fig:framework}):

\paragraph{Layer 1---Rules.}
The agent's distilled experience.
Each rule captures a specific pattern extracted from world feedback: a trigger condition over observable features and a corrective action.
Rules are \emph{deprecated, never deleted}---a deprecated rule stops firing but its knowledge is preserved.
This satisfies R3: the non-monotonic lifecycle.

Rules alone do not tell the agent which ones to trust.
A rule confirmed in three episodes but contradicted in five appears identical to one confirmed in five and contradicted in three if only a binary active/deprecated status is tracked.
This motivates the next layer.

\paragraph{Layer 2---Evidence.}
Each rule carries a persistent evidence log: a structured record of every episode where the rule was evaluated, whether it helped or hurt, and under what conditions.
Evidence accumulates across episodes and survives deprecation.
A rule contradicted once is noise; a rule contradicted consistently across varied conditions is a real failure.
This satisfies R1 (outcome-driven evaluation) and R2 (persistent, structured evidence).

Unlike Hindsight's scalar confidence scores~\citep{latimer2025hindsight}, our evidence logs preserve the full trail: which episodes, what conditions, how the rule affected reasoning.
When a rule is deprecated, the evidence explaining \emph{why} it was deprecated remains accessible---enabling the system to avoid re-proposing equivalent rules and to recognize when conditions change such that a deprecated rule might become relevant again.

\paragraph{Layer 3---Skills.}
The governance layer.
Skills read evidence across rules and control which rules occupy the agent's finite context, how to resolve conflicts, and when to abstain.
Skills evolve as evidence accumulates---early skills are tentative; later skills encode well-tested priority orderings and anti-patterns.
This satisfies R4: compositional governance.

The three-layer design means that skills are not arbitrary compositions but \emph{evidence-grounded} strategies.
A skill that prioritizes rule A over rule B does so because the evidence logs show A has been consistently reliable under the conditions where B fails.
This is motivated in part by the observation from SkillsBench~\citep{li2026skillsbench} that comprehensive, ungrounded skill sets can degrade performance; in our design, skills are focused by construction because they are derived from accumulated evidence rather than generated from scratch.

\subsection{The Curation Loop}
\label{sec:loop}

\begin{figure}[t]
  \centering
  \resizebox{\columnwidth}{!}{%
  \begin{tikzpicture}[
    node distance=0.6cm and 0.4cm,
    layer/.style={draw, rounded corners=4pt, minimum height=0.8cm, minimum width=2.8cm, font=\footnotesize, align=center, inner sep=5pt, line width=0.5pt},
    role/.style={draw, rounded corners=3pt, minimum height=0.65cm, minimum width=1.5cm, font=\scriptsize, align=center, inner sep=3pt, line width=0.4pt, dashed},
    arr/.style={-{Stealth[length=2.2mm]}, thick, color=black!70},
    dataarr/.style={-{Stealth[length=2mm]}, color=black!50, densely dashed},
    lbl/.style={font=\tiny\itshape, color=black!60},
  ]
    \node[layer, fill=blue!12] (rules) at (0,0) {\textbf{Layer 1: Rules} $\mathcal{L}$\\[-2pt]{\tiny trigger conditions + corrective actions}};
    \node[layer, fill=orange!12, below=0.5cm of rules] (evidence) {\textbf{Layer 2: Evidence} $\Xi$\\[-2pt]{\tiny per-rule logs: episode, outcome, conditions}};
    \node[layer, fill=green!12, below=0.5cm of evidence] (skills) {\textbf{Layer 3: Skills} $\mathcal{S}$\\[-2pt]{\tiny priority orderings, conflict resolution}};

    \draw[arr] (rules.south) -- node[right, lbl] {which to trust?} (evidence.north);
    \draw[arr] (evidence.south) -- node[right, lbl] {how to compose?} (skills.north);

    \node[role, fill=red!8] (critic) at ([xshift=3.5cm]rules.center) {Critic {\tiny (R1)}\\[-1pt]{\tiny evaluate vs.\ outcome}};
    \node[role, fill=yellow!15] (proposer) at ([xshift=3.5cm]evidence.center) {Proposer {\tiny (R2)}\\[-1pt]{\tiny append evidence,}\\[-1pt]{\tiny propose new rules}};
    \node[role, fill=purple!8] (curator) at ([xshift=3.5cm]skills.center) {Curator {\tiny (R3,R4)}\\[-1pt]{\tiny deprecate rules,}\\[-1pt]{\tiny evolve skills}};

    \node[font=\footnotesize\bfseries, above=0.4cm of critic] (world) {World Outcomes $y_t$};
    \draw[arr] (world) -- (critic);

    \draw[arr] (critic) -- (proposer);
    \draw[arr] (proposer) -- (curator);

    \draw[dataarr] (critic.west) -- (rules.east);
    \draw[dataarr] (proposer.west) -- (evidence.east);
    \draw[dataarr] (curator.west) -- (skills.east);

    \draw[arr, color=black!40] (curator.south) -- ++(0,-0.35) -| ([xshift=0.6cm]critic.east) -- (critic.east);

    \node[draw, rounded corners=3pt, fill=gray!10, minimum height=0.6cm, font=\scriptsize, align=center, inner sep=3pt, below=0.8cm of skills] (inference) {Inference: $\pi(a \mid x,\; \mathcal{L}_{\text{active}},\; \mathcal{S})$};
    \draw[arr] (skills.south) -- (inference.north);

  \end{tikzpicture}%
  }%
  \caption{Three-layer architecture with curation loop. \textbf{Left}: each layer solves a failure mode of the layer above---rules alone lack reliability signals; per-rule evidence alone lacks compositional reasoning; skills provide evidence-grounded governance. \textbf{Right}: the critic--proposer--curator pipeline connects world outcomes back to each layer. Evidence logs~$\Xi$ are append-only and persist across batches, grounding all governance decisions in observed outcomes. At inference, only active rules and current skills enter the agent's context.}
  \label{fig:framework}
\end{figure}
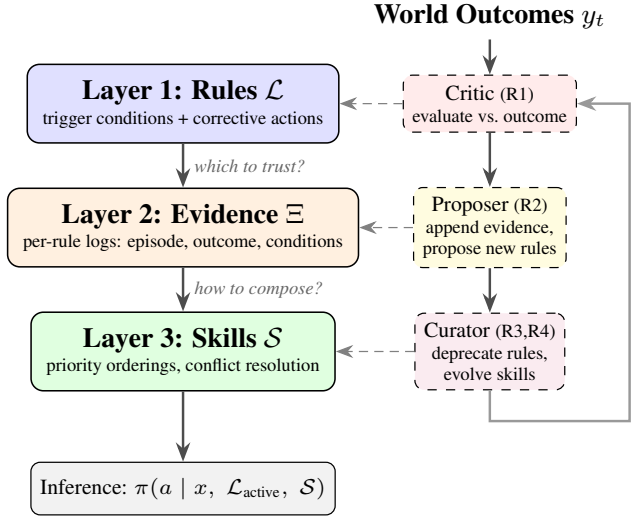

A critic--proposer--curator pipeline closes the feedback loop over batches of experience (\cref{fig:framework}).
Rather than detailing the mechanics of each role (see~\cref{sec:formal} for the formal description), we emphasize how the loop satisfies the requirements that existing approaches miss.

\paragraph{Closing the evaluation gap (R1).}
The critic compares rule-augmented reasoning against a zero-shot baseline on the same observed outcome.
This \emph{reasoning comparison} is strictly more informative than scalar success/failure: it attributes improvement or degradation to specific rules based on how they affected the agent's chain of thought, not just whether the final prediction was correct.
This is the evaluation mechanism that reflective accumulation~\citep{shinn2023reflexion} lacks entirely and that reflective refinement~\citep{zhao2023expel} approximates with scalar importance counts.

\paragraph{Preserving the evidence trail (R2).}
The proposer appends critic evaluations to per-rule evidence logs---it never modifies rules in place.
This append-only design is the key structural difference from prior work: when a rule is rewritten in ExpeL or TF-GRPO, all previously accumulated evidence is silently invalidated.
Our evidence logs survive rule deprecation, enabling the system to explain \emph{why} a rule was deprecated and to recognize when conditions change such that deprecated knowledge becomes relevant again.

\paragraph{Governing knowledge lifecycle (R3, R4).}
The curator reads cross-rule evidence to make lifecycle and governance decisions simultaneously: deprecating rules with consistently negative evidence (R3) and evolving skills that encode priority orderings, conflict resolution, and anti-patterns (R4).
Because both decisions are grounded in the same persistent evidence, they are mutually consistent---a skill never prioritizes a rule that the evidence contradicts.

\subsection{Formal Description}
\label{sec:formal}

Let $\mathcal{L}_k$, $\Xi_k$, and $\mathcal{S}_k$ denote the rule library, evidence logs, and skills after batch $k$.
Given a new batch of experience $B_k = \{(x_t, a_t, y_t)\}$:
\begin{align}
  E_k &= \textsc{Critic}(B_k, \mathcal{L}_{k-1}, \mathcal{S}_{k-1}) \label{eq:critic} \\
  \Xi_k, P_k &= \textsc{Proposer}(E_k, \mathcal{L}_{k-1}, \Xi_{k-1}) \label{eq:proposer} \\
  \mathcal{L}_k, \mathcal{S}_k &= \textsc{Curator}(\mathcal{L}_{k-1}, \mathcal{S}_{k-1}, \Xi_k, P_k) \label{eq:curator}
\end{align}
where $E_k$ is the set of verbal evaluations, $P_k$ is the set of newly proposed rules, and $\Xi_k$ extends the evidence logs with evaluations from batch $k$.
At inference, the agent executes $\pi(a \mid x, \mathcal{L}_K, \mathcal{S}_K)$---its parameters are frozen; only its context changes.

The key property of this loop is that $\Xi$ is append-only: evidence is never deleted or overwritten, even when the rule it concerns is deprecated.
This ensures that governance decisions (deprecation, skill evolution) are grounded in the complete history of observed outcomes.

\section{Empirical Validation}
\label{sec:experiments}

\subsection{Setup}
\label{sec:setup}

We use financial forecasting as our case study: world feedback (market returns) is abundant, objective, noisy, delayed, and non-stationary.
We adopt the same dataset and evaluation protocol as~\citet{cui2026hindsight}: daily OHLCV data for the top 5 S\&P~500 equities (AAPL, AMZN, FB, GOOGL, MSFT), 2013--2016 for learning and 2017 for testing, with 20-day candlestick chart inputs and 5-day prediction horizons.

\paragraph{What the agent learns.}
The base agent (Qwen3-VL-235B) observes a 20-day candlestick chart and produces a trading signal with directional prediction, scenario forecast, risk parameters, and chain-of-thought reasoning.
The critic, proposer, and curator use Claude Sonnet 4.6.
During the learning phase (2013--2016), the curation loop processes predictions in batches of ${\sim}$16 samples each.
After each batch, the critic evaluates predictions against realized market outcomes, producing verbal assessments that identify which rules helped, which hurt, and under what conditions.
The proposer appends these assessments to per-rule evidence logs and extracts new rules for uncovered error patterns---each rule is a natural-language statement specifying a trigger condition over observable chart features and a corrective action.
The curator reviews cross-rule evidence, deprecates rules with consistently negative evidence (removing them from the active context while preserving their evidence logs), and evolves skills---higher-order routing strategies that specify which rules to prioritize, how to resolve conflicts when multiple rules match, and when to abstain.
At inference (2017 test set), the agent receives only the active rules and current skills as additional context alongside the chart input; no model parameters are updated.

\paragraph{Baselines.}
All training-free methods use the same base agent and differ only in how accumulated experience is managed.
\emph{Zero-shot}: chart input only, no learned context.
\emph{Reflective Accumulation}~\citep{shinn2023reflexion,allard2026erl}: extracts rules from errors after each batch without curation---all rules are injected equally.
\emph{Reflective Refinement}~\citep{zhao2023expel,cai2025trainingfreegrpo}: extracts rules with importance scoring and in-place modification, but without persistent evidence logs or skill-based governance.
We report directional accuracy, scenario accuracy, average return per trade, Sharpe ratio, and maximum drawdown over 5 evaluation runs.

\subsection{Results}
\label{sec:results}

\begin{table*}[t]
  \caption{Prediction accuracy and trading performance on 2017 test set (mean $\pm$ std, 5 evaluation runs). All methods use the same base agent (Qwen3-VL-235B); they differ only in how accumulated experience is managed.}
  \label{tab:main}
  \begin{center}
    \begin{small}
        \begin{tabular}{lccccc}
          \toprule
          Method & Dir.\ Acc. & Scen.\ Acc. & Avg.\ Ret. & Sharpe & Max DD \\
          \midrule
          Zero-shot     & 51.2\%{\scriptsize$\pm$2.2}  & 23.8\%{\scriptsize$\pm$0.8}  & 0.16\%{\scriptsize$\pm$0.1}  & 0.53{\scriptsize$\pm$0.38}  & 34.5\%{\scriptsize$\pm$7.1}  \\
          Reflect.\ Accum.     & 46.3\%{\scriptsize$\pm$2.9}  & 23.3\%{\scriptsize$\pm$3.2}  & $-$0.08\%{\scriptsize$\pm$0.2}  & $-$0.12{\scriptsize$\pm$0.32}  & 35.2\%{\scriptsize$\pm$11.4}  \\
          Reflect.\ Refine.         & 51.1\%{\scriptsize$\pm$1.8}  & 24.7\%{\scriptsize$\pm$1.5}  & 0.14\%{\scriptsize$\pm$0.2}  & 0.36{\scriptsize$\pm$0.18}  & 24.4\%{\scriptsize$\pm$9.0}  \\
          \midrule
          Ours & \textbf{56.5\%}{\scriptsize$\pm$1.0}  & \textbf{29.0\%}{\scriptsize$\pm$2.5}  & \textbf{0.33\%}{\scriptsize$\pm$0.1}  & \textbf{1.00}{\scriptsize$\pm$0.3}  & \textbf{13.0\%}{\scriptsize$\pm$4.4}  \\
          \bottomrule
        \end{tabular}
    \end{small}
  \end{center}
  \vskip -0.1in
\end{table*}

The results (\cref{tab:main}) reveal three findings that directly map to the requirements framework (\cref{tab:gaps}):

\paragraph{Finding 1: No requirements $\rightarrow$ experience is harmful.}
Rules without evaluation, evidence, lifecycle, or governance degrade below zero-shot on every metric ($-4.9$pp accuracy, negative Sharpe).
Accumulated experience actively hurts.
This is the retention side of the dilemma: all rules are injected regardless of reliability, and whichever rule matches most closely dominates the agent's reasoning.

\paragraph{Finding 2: Partial requirements $\rightarrow$ partial recovery.}
Scalar evidence and in-place modification recover accuracy to near zero-shot levels but degrade risk-adjusted returns (Sharpe $0.36$ vs.\ $0.53$).
Better extraction helps, but without persistent evidence (R2) and governance (R4), conflicting rules still accumulate.
In our full loop, heavily cited rules that consistently produce wrong predictions are deprecated based on their accumulated negative evidence---the very frequency of citation becomes the signal for removal.
Without persistent evidence, in-place modification compounds the problem: when a rule is rewritten, all previously accumulated evidence is silently invalidated, and the system cannot distinguish a frequently wrong rule from a frequently useful one.

\paragraph{Finding 3: Full loop $\rightarrow$ genuine learning.}
Only when the full curation loop is active---addressing all four requirements---does the agent genuinely improve: $+5.3$pp directional accuracy, Sharpe nearly doubled, and maximum drawdown cut by $60\%$.
A key mechanism is that persistent evidence (R2) enables informed rule replacement (R3): new rules are not proposed solely from current-batch errors but are shaped by the accumulated evidence of \emph{why} predecessor rules failed, avoiding the same pitfalls.

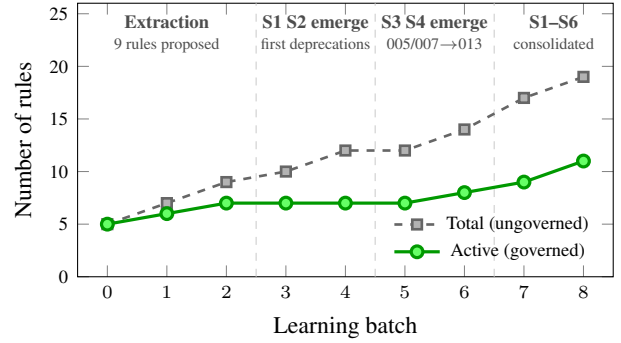
\begin{figure}[t]
  \centering
  \begin{tikzpicture}
    \begin{axis}[
      width=1.05\columnwidth,
      height=5.2cm,
      xlabel={Learning batch},
      ylabel={Number of rules},
      ymin=0, ymax=26, ytick={0,5,10,15,20,25}, yticklabels={0,5,10,15,20,25},
      xmin=-0.5, xmax=8.5,
      xtick={0,1,2,3,4,5,6,7,8},
      legend style={at={(0.98,0.02)}, anchor=south east, font=\scriptsize, draw=none, fill=none},
      tick label style={font=\scriptsize},
      label style={font=\small},
      every axis plot/.append style={thick},
      set layers,
    ]
    \draw[thin, black!20, dashed] (axis cs:2.5,0) -- (axis cs:2.5,26);
    \draw[thin, black!20, dashed] (axis cs:4.5,0) -- (axis cs:4.5,26);
    \draw[thin, black!20, dashed] (axis cs:6.5,0) -- (axis cs:6.5,26);

    \node[font=\scriptsize, text=black!70, align=center, anchor=north] at (axis cs:1,25.8)
      {\textbf{Extraction}\\{\tiny 9 rules proposed}};
    \node[font=\scriptsize, text=black!70, align=center, anchor=north] at (axis cs:3.5,25.8)
      {\textbf{S1 S2 emerge}\\{\tiny first deprecations}};
    \node[font=\scriptsize, text=black!70, align=center, anchor=north] at (axis cs:5.5,25.8)
      {\textbf{S3 S4 emerge}\\{\tiny 005/007$\to$013}};
    \node[font=\scriptsize, text=black!70, align=center, anchor=north] at (axis cs:7.5,25.8)
      {\textbf{S1--S6}\\{\tiny consolidated}};

    \addplot[dashed, color=black!60, mark=square*, mark size=1.8pt, mark options={solid, fill=black!30}, line width=1pt]
      coordinates {(0,5) (1,7) (2,9) (3,10) (4,12) (5,12) (6,14) (7,17) (8,19)};
    \addplot[solid, color=green!60!black, mark=*, mark size=2.2pt, mark options={solid, fill=green!60}, line width=1.2pt]
      coordinates {(0,5) (1,6) (2,7) (3,7) (4,7) (5,7) (6,8) (7,9) (8,11)};
    \legend{Total (ungoverned), Active (governed)}

    \end{axis}
  \end{tikzpicture}
  \caption{Rule library evolution. Dashed: cumulative rules (ungoverned). Solid: active rules after deprecation (governed). The widening gap is the governance contribution. Background bands mark curation phases; skill themes are detailed in \cref{tab:skills}.}
  \label{fig:evolution}
\end{figure}

\begin{table}[t]
  \caption{Learned skill themes and their provenance. Each skill synthesizes evidence across batches, governing active rules and learning from deprecated ones (R2--R4 in practice).}
  \label{tab:skills}
  \begin{center}
    \resizebox{\columnwidth}{!}{%
    \begin{small}
        \begin{tabular}{llll}
          \toprule
          Skill Theme & Active Rules & Deprecated & Evidence \\
          \midrule
          S1: Capitulation reversal & rule\_002, 010 & --- & Batches 4,6,7 \\
          S2: Post-decline recovery & rule\_008 & --- & Batches 3,6,7 \\
          S3: Absorbed distribution & rule\_013, 016 & rule\_005, 007 & Batches 6,7 \\
          S4: Distribution staircase & rule\_006, 015 & --- & Batches 4,5,6 \\
          S5: Post-decline stabilization & rule\_014 & --- & Batch 7 \\
          S6: Trend exhaustion & rule\_017 & --- & Batch 7 \\
          \bottomrule
        \end{tabular}
    \end{small}
    }%
  \end{center}
  \vskip -0.1in
\end{table}

\section{Discussion}
\label{sec:discussion}

\paragraph{From memory management to insight governance.}
Modern agent memory architectures have independently converged on storage layers that resemble our three layers: declarative knowledge (rules), session histories and belief tracking (evidence), and procedural skills~\citep{latimer2025hindsight,packer2023memgpt,hu2025memory}.
Hindsight's retain-recall-reflect paradigm~\citep{latimer2025hindsight} provides the right structure for memory management; our work suggests that learning from non-stationary world feedback requires extending \emph{reflect} with outcome-driven evaluation, persistent evidence, and compositional governance.
The extraction-to-governance gap we identify is not unique to verbal RL---any system where agents accumulate experience from world outcomes faces the same challenge.
The four requirements may serve as a design checklist: any such agent should ask whether its insights are evaluated (R1), evidenced (R2), lifecycle-managed (R3), and governed (R4).

\paragraph{Toward heterogeneous world feedback.}
Real-world feedback signals vary widely in noise, delay, density, and stationarity---from immediate binary task success to delayed continuous market returns to dense robotic control signals.
Our curation loop is validated on one point in this space (noisy, delayed, non-stationary financial outcomes).
Whether the same evidence structures and governance patterns transfer across feedback types---and systematic evaluation across the heterogeneous feedback landscape the workshop envisions---remains open.

\paragraph{Meta-curation: learning to govern.}
Our curation mechanism---critic criteria, deprecation thresholds, skill structure---is hand-designed.
Can world feedback also drive the evolution of the curation mechanism itself?
Recent work on metacognitive self-modification~\citep{zhang2026hyperagents} demonstrates that agents can learn to modify their own reasoning strategies.
Applying this to the curation loop would close a second feedback loop: world outcomes improving not just what the agent knows, but how the agent manages what it knows---a natural extension from learning \emph{with} world feedback to learning \emph{how to learn from} world feedback.

\paragraph{Limitations.}
Our empirical validation covers a single domain (S\&P~500 equities, 2013--2017) in a predominantly bullish regime.
Evidence logs are natural language, growing linearly with episodes; structured or embedding-based representations may scale better while preserving auditability.
Evidence is currently treated uniformly across time; temporal discounting could improve adaptation speed in rapidly shifting environments but risks forgetting lessons from rare conditions.

\section{Conclusion}
\label{sec:conclusion}

Agents that learn from world feedback face a retention-forgetting dilemma that no single layer of memory can resolve.
We identified four requirements for navigating this dilemma---outcome-driven evaluation, persistent structured evidence, non-monotonic knowledge lifecycle, and compositional governance---and showed, through examination of representative training-free verbal learning methods, that no existing approach satisfies all four.

Our three-layer architecture---rules, evidence, and skills connected by a feedback-driven curation loop---satisfies all four requirements.
On financial forecasting as a case study, the same accumulated experience either degrades performance ($-4.9$pp accuracy, negative Sharpe) or dramatically improves it ($+5.3$pp accuracy, $2\times$ Sharpe, $60\%$ less drawdown), depending on whether the curation loop is present.

The same experience, the same agent, the same three layers: only the curation mechanism determines whether the agent improves or degrades.
In this setting, the primary bottleneck is not experience extraction but insight governance---a framing we offer as a working hypothesis for agents learning from world feedback.

\section*{Impact Statement}

This paper addresses the general problem of how LLM agents learn from world feedback, validated on financial forecasting as a case study.
The learned rules and skills are human-readable and auditable, supporting transparency in AI-assisted decision-making.
We caution that our empirical results are based on historical financial data in a predominantly bullish regime (2013--2017 S\&P 500) and should not be interpreted as evidence of live trading viability.
The four requirements and architectural principles we identify are intended to inform the design of agent learning systems, not to provide investment advice.


\bibliography{references}
\bibliographystyle{icml2026}

\end{document}